\documentclass[10pt,a4paper]{report}
\usepackage{amsmath}
\usepackage{amsfonts}
\usepackage{amssymb}
\usepackage{mathtools}
\usepackage[colorlinks,linkcolor=blue]{hyperref}
\usepackage[nameinlink,noabbrev]{cleveref}
\usepackage{tikz}
\usetikzlibrary{arrows}
\usepackage{arxiv}
\usepackage[numbers]{natbib}
\usepackage{setspace}
\author{Amelia E. Pollard\\Department of Computer Science\\University of Manchester\\\texttt{amelia.pollard@postgrad.manchester.ac.uk} \and Jonathan L. Shapiro\\Department of Computer Science\\University of Manchester\\\texttt{jonathan.shapiro@manchester.ac.uk}}
\title{Eliminating Catastrophic Interference with Biased Competition}
\begin{document}
\maketitle

\begin{abstract}
We present here a model to take advantage of the multi-task nature of complex datasets by learning to separate tasks and subtasks in and end to end manner by biasing competitive interactions in the network. This method does not require additional labelling or reformatting of data in a dataset. We propose an alternate view to the monolithic one-task-fits-all learning of multi-task problems, and describe a model based on a theory of neuronal attention from neuroscience, proposed by \citet{desimone1998visual}. We create and exhibit a new toy dataset, based on the MNIST\citep{deng2012mnist} dataset, which we call MNIST-QA, for testing Visual Question Answering architectures in a low-dimensional environment while preserving the more difficult components of the Visual Question Answering task, and demonstrate the proposed network architecture on this new dataset, as well as on COCO-QA\citep{ren2015exploring}  and DAQUAR-FULL\citep{malinowski2015ask}. We then demonstrate that this model eliminates catastrophic interference between tasks on a newly created toy dataset and provides competitive results in the Visual Question Answering space.  We provide further evidence that Visual Question Answering can be approached as a multi-task problem, and demonstrate that this new architecture based on the Biased Competition model\citep{desimone1998visual} is capable of learning to separate and learn the tasks in an end-to-end fashion without the need for task labels.

\end{abstract}
\section{Introduction}
Deep neural networks of ever increasing complexity and scope have been developed focusing on tasks previously only achievable by humans enabled by the rise in accessibility of GPU computing resources previously reserved for only the largest super-computers. Visual Question Answering in particular represents this trend more than most, combining image processing, natural language processing, and machine reasoning. Our previous research indicated that Visual Question Answering can be tackled as a multi-task learning problem\citep{Pollard2020}, where each question type can be considered a separate task.
Visual Question Answering(VQA) is the process of showing a machine learning system a natural scene and posing a question about that scene in natural language. First demonstrated on a large complex dataset in \citet{antol2015vqa}, the problem is a difficult one combining computer vision and natural language processing with spatial and rational reasoning.
Approaches to the Visual Question Answering problem have fallen into four distinct categories: joint-embedding, attention models, knowledge-based models, and compositional models. The first category, joint-embedding, takes features extracted from both the input image and question and combines them in one of a variety of multimodal methods to generate a shared representation of both the image and question, from which an answer can be extracted. An example of this approach is Malinowski et al.\citep{malinowski2014nips} where a CNN extracts image features before concatenating them with embedded question words and feeding them into an LSTM.\\
The second category, attention models, has received the most focus in research in recent years. This approach has been successful in several fields, from computer vision\citep{mnih2014recurrent} to speech recognition\citep{chorowski2015attention}. Examples include the Stacked Attention Network \citep{yang2016stacked}, LSTM-Attention\citep{zhu2016visual7w} and many others. These approaches have thus far taken an approach to attention focusing on emulating the human eye, with certain spatial areas of the images being brought into focus while others are consigned to the background.
The third category of approaches to Visual Question Answering, knowledge based, relies on external sources of knowledge which may or may not be available during training. An excellent example of this type of network can be found in \citet{wu2016ask}, where a combination of caption generation and visual attention is used to query a database of knowledge, which in turn is used to inform the Visual Question Answering network.\\
Finally, we come to the fourth class of Visual Question Answering networks: the compositional model. This type of model uses the question features to generate a network or program capable of answering the question. Examples of this type include \citet{andreas2016neural} and \citet{johnson2017inferring}. Andreas et al. select neural network modules to generate reasonable answers through semantic parsing of the question, whereas Johnson et al. use a Seq2Seq\citep{sutskever2014sequence} LSTM model to generate "programs" which are then executed to generate answers.

In this work, the model we propose combines attentional and compositional approaches, with attention not focused on the image, but on the network itself. As shown in \citet{andreas2016neural}, each question in a Visual Question Answering dataset can be broken down into a series of subtasks which can be tackled effectively to produce high quality answers. We propose that this series of Visual Question Answering subtasks is analogous to a multi-task environment, as shown to be in \citet{Pollard2020}. As such, we hypothesise that we can take advantage of the effects of multi-task learning to improve the performance of Visual Question Answering tasks. Unlike the method proposed in Neural Module Networks\citep{andreas2016neural} however, the model we propose is capable of learning to separate those tasks during training and run time in an unsupervised manner by learning to enable and disable portions of the network, essentially creating adaptable neural modules. Our contribution can therefore be considered as a more general form of a neural module network, with a specific focus on the formation of subnetworks to tackle inter-task interference in an unsupervised manner. 

We hypothesise that this inter-task interference between question types is in fact the catastrophic interference typically observed when training a network on multiple distinct tasks sequentially, as described in \citet{french1999catastrophic}. We believe this hypothesis holds merit because while the questions of the Visual Question Answering datasets clearly share a domain, they fall into distinct categories which, in other works, would constitute a whole task in and of themselves. As such, training a Visual Question Answering dataset shares some notable similarities with training a neural network in a multi-task environment, as evidenced by the performance increases observed by training a VQA dataset in a multi-task architecture in \citet{Pollard2020}.

\subsection{Multi-task Learning}
Multi-task learning\citep{caruana1997multitask} is the paradigm of learning distinct tasks within the same domain. The core principle of multi-task learning is that a network can be trained over a primary task, with one or more additional secondary tasks also being trained in order to improve the generalisation of the domain representation. The search for a parameter set in the hypothesis space which approximates all of the learned tasks simultaneously leads to a transfer of information between tasks, strengthening each one\citep{baxter1997bayesian}.

In this context, we define a domain $D$ as a combination of a feature space $X$ and a distribution $P(y|X)$, where a task $T$ is defined as a label space $Y$ and a predictive function $f(.)$ which is learned from training data and can be represented as a conditional probability $P(y|x)$, which is to say, the probability of observing label $y \in Y$ from the domain $D$. In the case of multi-task datasets, the domain remains consistent, but the tasks are distinct, such that $P(y|x)$ is distinct between each task.\\

In the ideal case, for each task in a domain, one would create a network to approximate the function $P(y|x)$ from training data with a perfect internal representation of the domain. Unfortunately, not only is this resource intensive - requiring a significant amount of compute power for each task - it is also rarely possible in cases where individual tasks are not labelled and thus easily separable from one another. Furthermore it is beneficial to learn a shared representation where tasks share some commonalities (e.g. identifying dogs and cats in natural scenes) as this causes a network to learn a more complete representation of the domain. 

While a multi-task approach has been used to augment networks in a number of domains, it requires either the creation of additional labelled data or the reformatting of existing data in order to achieve this more general representation. In this paper, the technique we propose allows a Visual Question Answering network to benefit from the effects of multi-task learning without those constraints, instead allowing the whole dataset to be used as-is with no additional formatting or task-type labels required. The network architecture proposed takes inspiration from the theory of Biased Competition\cite{desimone1998visual} (discussed below) and is designed to allow for the automatic separation of the network into a collection of subnetworks which can be activated at any time. This, we hypothesise, allows the network to learn to separate tasks and learn them apart from one another thus preventing the catastrophic interference typically observed.

\subsection{Biased Competition}

The theory of Biased Competition originates from the study of attention in neuroscience. The general principle of this theory is that neurons in the visual cortex are suppressed or excited by feedback from spatial and object aware components of the brain, which are informed by working memory. This is achieved through a process of competitive interactions between neurons, with some neurons being excited by the feedback signal, introducing a bias.\\

In his seminal paper on the topic, Desimone\cite{desimone1998visual} argues that objects in the visual field compete for processing in the mammalian brain. As a consequence of this competition for processing, a system evolved to prioritise the objects and features in the visual field that contain behaviourally relevant information. The core of the Biased Competition theory is that a combination of top-down and bottom-up processes result in biasing the visual cortex towards this behaviourally relevant data. 

Desimone further describes five central tenets of his proposed model:
\begin{itemize}
\item{Objects in the visual field compete for neural resources. That is to say, two or more objects presented simultaneously will be in competition with one another for attention.}
\item{The strongest competitive interactions are achieved when the stimuli are activating similar neurons, for example, two faces would compete with each other for attention more than a face and a car and two objects very close to one another would compete more than two objects very far apart}
\item{Competing visual stimuli are biased towards one or the other by many mechanisms, rather than a single overall attention control system. These mechanisms include bottom-up mechanisms such as one object having more visual contrast than another, as well as top-down mechanisms, such as the biasing result of searching for one's car}
\item{The feedback bias is not purely spatial. That is to say, the biasing input can prime neurons for textures, colours, shapes, and any other feature that is behaviourally relevant}
\item{Lastly the most speculative tenet as described by Desimone, the top-down biasing input is generated from the working memory sections of the brain - the parts responsible for task-driven behaviour}
\end{itemize}

There is considerable evidence for the theory of Biased Competition in the human brain, and a great many experiments have been performed to analyse the effects that it can have on human visual attentional systems. 

In this paper, we take a considerably simplified view of the Biased Competition model, whereby neurons are entirely suppressed by the action of another neural network which is informed by context - in this case, the question vector. \\

The resulting method is designed to allow for the automatic separation of the network into a collection of subnetworks which can be activated at any time. This, we hypothesise, will allow the network to learn distinct tasks and to learn these tasks separately from one another, thus preventing the catastrophic interference typically observed. Thus, the primary contribution of this paper is a technique which enables neural networks to learn several tasks concurrently without the typically observed destructive interference between those tasks, in a manner which does not require additional or reformatted data. 

The primary contribution of this paper is a technique by which neural networks are able to learn several tasks concurrently without the typically observed destructive interference between those tasks in a manner which does not require additional or reformatted data. Subnetwork formation of locally competitive models has already been explored \citep{srivastava2014understanding}, however our approach differs by the application of a top-down biasing input, which biases those competitive interactions towards more behaviourally relevant selections, in line with the theory of Biased Competition as proposed by Desimone. In order to replicate the top-down biasing functionality of the Biased Competition system, the model must receive some input from the scene, and some input which provides context, before generating a signal which biases a competitive network which performs the actual classification.\\

It has been shown that tasks which share a representation can augment one another when learned using a method which allows that shared representation to be fully utilised. Caruna\citep{caruana1997multitask} demonstrated that networks with multiple outputs and a shared hidden layer can be trained by backpropagation to learn this representation, however this approach requires that the tasks are distinctly labelled in the training data. The approach proposed here does not require such a constraint, as a section of the network learns to split tasks and optimise the shared representation using only the error signal from the final prediction. This allows datasets which were not previously considered "multi-task" datasets to be optimised in a way which allows this shared representation to form optimally. This can greatly augment the performance of the final predictive model.

Thus we differ from the standard approach of multi-task networks where the goal is to solve multiple tasks on one input, for example: when a network is shown a picture of a flower, and must determine the colour, the shape and the texture simultaneously. This causes the formation of a more complete internal representation of the domain and enables the individual tasks to perform more effectively. In this work we take advantage of the more general internal representation, though solve only one question about the input, which is itself taken as input to the network.

We propose here a method by which one network is capable of learning multiple distributions that do not interfere with one another, and indeed in certain cases, actually augment one another. We will demonstrate this method by showing its effect on Visual Question Answering problems, which we have previously demonstrated can be augmented by being treating as a multi-task problem.\\

Due to the extremely high-dimensional nature of the Visual Question Answering problem, we also developed a simplistic dataset based on the MNIST numbers and auto-generated questions with the goal of providing a less complex dataset with which we use to analyse the neural modules created by the architecture. We then demonstrated this technique on both COCO-QA and DAQUAR datasets, and compared these results against a previous state-of-the-art joint-embedding model.

\section{Methodology}
We hypothesised that applying our Biased Competition model to a network would allow the network to form distinct subnetworks, analogous to neural modules, which would be capable of learning separate tasks concurrently without causing destructive interference between them. In order to test this hypothesis, we built a model of the Biased Competition network and measured the formation of subnetworks and the reduction in interference as tasks were introduced. We measured interference by comparing performance against expected performance while training over the combination of question types in the Visual Question Answering datasets. We make the assumption that where interference is minimised the tasks will become independent of one another. The formation of subnetworks was measured by performing a clustering over the binary matrices produced by the biasing input, with the assumption that these matrices will be measurably distinct from one another as a result of the subnetwork formation.

\subsection{Our Biased Competition Model}
In this section, we set out the formal definition of our model, and the proposed alternate view of multi-task learning as a collection of subtasks.
Previous work in multi-task learning has been based on the premise that for a set of tasks $T$ there is a corresponding set of functions which describes each of those tasks. It has been further proposed that in this case, the best way to solve for all of them is to find a function $f(X)$ which best approximates all of them. That is to say, for all tasks in a domain there exists a single function which can approximate the relationship between data and labels. This can be formally expressed as follows:
\begin{equation}
\forall t \in T \exists f(X) \approx t
\end{equation}
We instead propose that a better way to view this set of problems is to treat each task in a domain as a collection of subtasks which may be shared between other tasks in the domain. Thus, we propose that the approximator function $f_t$ for the task $t$ can be treated as if it were a summation of several subtasks $f_t^n$. As such, we formally define the approximator for the task $t$ in the domain $T$ as:
\begin{equation}
f_t = f_t^1 + f_t^2 ... f_t^n
\end{equation}
In addition, we propose that some subtasks are common to several of the tasks in $T$, such that for some tasks, $f_t^n$ may be shared between those tasks.

Therefore, instead of finding $f(X)$ by searching for an overall function which approximates $T$, we endeavour to find the summation of $f_t$ over $t \in T$, and as a result of the shared subtasks, we can best approximate it by finding a function which is parameterize by $t$ and chooses the combination of appropriate subtasks based on this. Thus, we define a function $g(t)$ which acts as a selector for the learned subtasks:
\begin{equation}
f(X) = g(t,f_t^1)f_t^1 + g(t,f_t^2)f_t^2 ... g(t,f_t^n)f_t^n
\end{equation}
where $g(t)$ is a function which generates a $0$ or $1$ based on $t$.

We therefore solve the multi-task problem as a summation of several subtasks selected from a pool of possible subtasks resulting in a more accurate approximation of each $f_t$ within $f(X)$. As a result of this approach, we can take advantage of the nature of the back-propagation of errors algorithm, which allows us to ignore components with zero input, thus enabling the network to avoid the issue of destructive interference during training for the individual subtasks which are not being used for any given example.\\

Applying this hypothesis to the problem of Visual Question Answering, we describe the model as follows: Let $Q$ be the question representation vector, and $V_0$ be the image representation matrix. $V_1$ is the resulting image representation matrix with the irrelevant parts of the image set to $0$. $C$ is a hyperparameter representing the control strength of the bias function, where $C=0$ is essentially equivalent to random Dropout and $C=1$ results in the biasing signal being entirely dependent on the question representation vector.
\begin{equation}
V_1 = V_0 \odot \lfloor (\sigma(W_Q Q + b_Q))*C + \mathcal{U}[0,1)*(1 - C) +0.5\rfloor
\end{equation}
Where the control strength parameter is $1$, this is effectively 
\begin{equation}
V_1 = V_0 \odot {\rm I\!P(\sigma(W_Q Q + b_Q))}
\end{equation}
Where
\[
\begin{array}{@{} r @{} c @{} l @{} }
${\rm I\!P(x)} =$
\begin{cases}
1 | x  \\
0 |(1-x)
\end{cases}
\end{array}
\]
or, less formally, we sample from the discrete distribution ${\rm I\!P(x)}$ where $x$ indicates the probability of observing a $1$.\\
Note that $\lfloor x \rfloor$ indicates the floor function, $\odot$ indicates element-wise matrix multiplication, $\mathcal{U}[0,1)$ indicates sampling from a uniform distribution bounded between $0$ and $1$, inclusive of $0$ and exclusive of $1$.\\
Being a step function, $\lfloor x \rfloor$ is not continuously differentiable, and so for the purposes of training, we use the more abstract interpretation as a discrete probability distribution and differentiate the expected value as $\langle{\rm I\!P(x)}\rangle = x$ and therefore $\frac{d}{dx}{\rm I\!P(x)} = 1$.\\
Finally, to emulate competition between neurons, we normalise $V_1$. This gives the following final Bias function as:
\begin{equation}
V_1 = ||V_0 \odot \lfloor \sigma(W_Q Q + b_Q)*C + \mathcal{U}[0,1)*(1 - C) +0.5 \rfloor||
\end{equation}
The network architecture is designed to emulate the rudimentary structure of the method used by the model of Biased Competition proposed by Desminone\citep{desimone1998visual}. Using a recurrent network of memory informed by visual data, which in turn informs the biasing input, thereby modifying the state of the network, which again informs the memory. Specifically, we use the pre-trained ImageNet CNN (VGG19\citep{simonyan2014very}) in addition to a bag-of-words CNN model for the natural language processing of the questions. As \citet{yang2016stacked} argued in their paper "Stacked Attention Networks", we take the view that Visual Question Answering is a multi-step process, and so in combination with the multi-step nature of the Biased Competition model proposed by Desimone, we repeat the biasing layer several times as indicated below by the looped arrows in the diagram below.
\begin{center}
\begin{tikzpicture}

\tikzset{vertex/.style = {shape=rectangle,draw,minimum size=1.5em, fill=blue!20}}
\tikzset{biasvertex/.style = {shape=rectangle,draw,minimum size=1.5em, fill=green!20}}
\tikzset{edge/.style = {->,> = latex}}
\node[vertex,anchor=west] (imagein) at  (0,0) {$x_I$};
\node[vertex,anchor=west] (questionin) at  (0,3) {$x_Q$};
\node[vertex,anchor=west] (questioncnn) at  (4.25,3) {NLP CNN};
\node[vertex,anchor=west] (imagecnn) at  (1.5,0) {ImageNet CNN};
\node[vertex,anchor=west] (lstm) at  (4.25,1.5) {LSTM};
\node[vertex,anchor=west] (embedding) at (1.5,3) {embedding};
\node[biasvertex,anchor=west] (bias) at (4.5,0) {Bias};
\node[vertex,anchor=west] (class) at (6,0) {classification MLP};
\node[vertex,anchor=west] (yhat) at (9.5,0) {$\hat{y}$};

\draw[edge] (imagein) to (imagecnn);
\draw[edge] (imagecnn) to (bias);
\draw[edge] (bias) to[bend left] (lstm);
\draw[edge] (lstm) to[bend left] (bias);
\draw[edge] (questionin) to (embedding);
\draw[edge] (embedding) to (questioncnn);
\draw[edge] (questioncnn) to (lstm);
\draw[edge] (bias) to (class);
\draw[edge] (class) to (yhat);

\end{tikzpicture}
\end{center}

\subsection{Further Details}
Using the context of the question representation vector, we generate biasing input which inhibits parts of the network and allows specific parts of the network to be trained for specific contexts. For example, a question regarding the colour of an object would suppress the areas of the network responsible for object classification and excite those responsible for colour classification. The purpose of this is to reduce the amount of unneeded information, which acts essentially as noise, when performing a task.

The proposed model of attention centres around an ability to process context into an internal representation, and then apply that representation to a scene to suppress some elements of the scene while accentuating others. This functionality essentially allows the network to "switch off" unneeded neurons, reducing their impact on the final task output. This reduced level of output, in combination with a standard back-propagation algorithm, causes the weights of unused neurons not to be adjusted - allowing for training of only the relationship between the target object and the corresponding output. 
In the case where multiple tasks are present in the target domain, we hypothesise that the network will learn to separate tasks into "subnetworks", that is, by disabling sections of the network for particular tasks and not for others, the network is able to learn multiple task distributions that do not interfere with one another. This contrasts with other methods of multi-task learning, which seek to learn a shared representation of all tasks, in an attempt to find a parameter set which approximates all of the component tasks together.
\section{Experiments}
To demonstrate the network's ability to reduce the effect of catastrophic interference in multi-task datasets, we first needed to use a dataset which was easily formatted into context and environment, and which could also be classified into easily separable classes of problem. As these requirements naturally lend themselves to a Visual Question Answering problem set, we first created a simplified VQA dataset in which categories were easily separable, before testing on a more established VQA dataset which was more difficult to categorise.  
\subsection{MNIST-QA Dataset}
The designed dataset falls into two parts: the images and the questions about those images.
Similar datasets, such as DAQUAR and COCO-QA include questions about colour, object, and count. However, as we are using the MNIST dataset as a source for the images which is monochromatic, colour questions were not included in this dataset. The DAQUAR dataset also includes relative position questions. As these are the more difficult part of the task, they were also included in this dataset in order to properly represent the complexities of the original problem space. Our dataset therefore includes questions about objects, counting, relative positions, and relative counting. The varied question types in our dataset are designed to force a scenario where multiple problems use the same elements of other problems. For example, in a relative position object question, and a relative position count question, both tasks require an understanding of relative position. If our hypothesis holds, we would expect to see a reduction in interference between both tasks.\\

The specification of the proposed dataset is as follows:\\
\begin{itemize}

\item{The images consist of a 3 by 3 grid of either blank spaces or samples from the MNIST dataset forming a 9-slot grid.}
\item{The questions will fall into four classes: Object, Count, Relative Object, Relative Count.}
\item{Object questions will consist of simple position and object questions, such as "What is the number in the top right corner?"}
\item{Count questions will consist of simple position and count questions, such as "How many numbers are in the second row?"}
\item{Relative object questions will consist of object questions with relative positions, such as "What is the number to the right of the 3?"}
\item{Relative count questions will consist of count questions with relative positions, such as
"How many numbers are in the row below the row that contains the 3?"}

\end{itemize}

The images themselves are composed of 9 patches of 28x28 images in a square grid (from either the MNIST dataset or blank squares), giving a total dimensionality of $d=7056$ for the image. This needed to be processed by a CNN in order to be reduced to a usable dimensionality for classification.\\
For clarity, we will now refer to this dataset as the MNIST-QA dataset.
\subsection{Measurements}
Our hypothesis indicates two testable measures; the first measure is inter-task interference, where showing that the tasks learned together perform as well as or better than the tasks learned separately. The second measure is to demonstrate the mechanism by which this occurs, and to show that the network learns to separate tasks by type, without type labels training it to do so.

To test these measures, we first required the ability to label task types. While DAQUAR does not provide question type classification, we were able to perform a naive classification by searching for relevant keywords. For example, questions containing "colour" or "red" are classified as colour questions, while questions containing "where" or "behind" are position questions. Some questions fall into both categories, such as "Where is the red object?".
The COCOQA dataset comes pre-labelled with question types (object, number, colour, location), and we were able to use these directly. This was also true for the MNIST-QA dataset detailed in section 0.5.1.
As we were able to classify the questions by type in the datasets and measure individual accuracy on each possible combination of question types, we were able to demonstrate whether interference was happening between each question type, and which tasks interfere with each other. For the purposes of eliminating other possible sources of performance changes, we balanced the number of question types against the number of samples shown to the network. We did this by keeping batch size consistent, while varying the number of epochs to preserve an equal number of samples for each question type the network is trained on.\\
In order to show that the individual question tasks are being separated correctly, we validated the above experiment by assigning random labels to the individual tasks and measuring the observed accuracy. If our hypothesis holds, we should not observe the same interference effects with the random task type labels. \\

To further test the hypothesis, we collected the binary masks which control neuron activation and performed a clustering with those generated by the training set. Using these clusters, we were able to predict the question type purely by observing the cluster it belongs to. We assumed the produced masks already have structure that clusters them, and we analysed the structure to evidence this. It is of note that the biasing input was not explicitly trained to distinguish question types, or indeed to attain any goal other than decreasing the loss function of the whole network, thus the presence of strongly differentiable clusters indicates the network is learning to separate tasks as predicted. When combined with evidence of reduced interference between tasks as shown by the increased independence between errors per class, this strongly indicates that our hypothesis is correct. 

To evaluate these clusters, we learn the clusters using the k-means clustering algorithm and selected the best value of   $k$ and centroid starting points by repeated testing over a variety of values (for $k$, values of 1 to 20 were tested) and selecting the values with the best validation score on the test set. Once the clusters were defined, we measured entropy, purity, and recall before collating them into a final F-Measure score. We take the null hypothesis as being that any clusters formed bear no relation to the question types, and tested this by assigning random labels to the clusters, on the assumption that clusters which were not strongly formed would have significantly lower predictive ability for the class labels than those which were strongly formed.

In addition to the cluster analysis representing separation of tasks, we also wished to show that the transfer learning between tasks could augment individual performance of each task. We did this by training the network on all combinations of tasks, with the assumption that tasks which were independent of one another would result in a final accuracy which was equal to the average of all component tasks together. We predicted that tasks which transfer knowledge between each other would perform better than the average of the component tasks trained alone, while tasks which interfere with one another would perform worse that the average of the tasks alone.
\\
All networks were trained until convergence with Nadam\citep{dozat2016incorporating}, before being fine-tuned by SGD with momentum, as described in \citet{keskar2017improving}.

\section{Results}
Below we present the results of running the proposed Biased Competition Network against the MNIST-QA dataset. Each line of the table represents a possible combination of tasks in the MNIST-QA network, the accuracy attained, the expected accuracy (as the average of the accuracy of the component tasks), and the difference between the expected and actual accuracy. A checkmark(\checkmark) in the columns for count, number, relative count, and relative number (c, n, rc, rn) show the presence of those question types in the trained dataset. Where there is only one component, the expected accuracy is the attained accuracy, and so the difference is always 0.

\begin{table}[h]
\small
\centering
\caption{Biased Competition Network Results by Question Type on MNIST-QA. Column names are abbreviated as c - count, n - number, rc - relative position count, rn - relative position number. Expected accuracy is calculated by the weighted average accuracy of the component tasks.}
\label{biastable}
\begin{tabular}{|l|l|l|l|l|l|l|l|l|l|}
\hline
\# of qtypes & Accuracy & c & n & rc & rn & Expected Accuracy & Difference    \\ \hline
1            & 99.80    & \checkmark     &        &                &                 & 99.80              & 0.00\\\hline
1            & 96.60    &       & \checkmark       &                &                 & 96.60              & 0.00\\ \hline
2            & 98.52    & \checkmark      & \checkmark       &                &                 & 98.20              & 0.32        \\ \hline
1            & 70.00    &       &        & \checkmark               &                 & 70.00              & 0.00            \\ \hline
2            & 88.06    & \checkmark      &        & \checkmark               &                 & 84.90              & 3.16        \\ \hline
2            & 83.62    &       & \checkmark       & \checkmark               &                 & 83.30              & 0.32         \\ \hline
3            & 90.93    & \checkmark      & \checkmark       & \checkmark               &                 & 88.80              & 2.13       \\ \hline
1            & 46.00    &       &        &                & \checkmark                & 46.00              & 0.00            \\ \hline
2            & 73.34    & \checkmark      &        &                & \checkmark                & 72.90              & 0.44         \\ \hline
2            & 71.30    &       & \checkmark       &                & \checkmark                & 71.30              & 0.00          \\ \hline
3            & 82.64    & \checkmark      & \checkmark       &                & \checkmark                & 80.80              & 1.84       \\ \hline
2            & 58.00    &       &        & \checkmark               & \checkmark                & 58.00              & 0.00          \\ \hline
3            & 72.79    & \checkmark      &        & \checkmark               & \checkmark                & 71.93              & 0.86 \\ \hline
3            & 70.81    &       & \checkmark       & \checkmark               & \checkmark                & 70.87              & -0.06 \\ \hline
4            & 80.73    & \checkmark      & \checkmark       & \checkmark               & \checkmark                & 78.10              & 2.63        \\ \hline
\end{tabular}
\end{table}
We found that with the Biased Competition network, tasks are prevented from negatively interfering with one another, and compared this to an equivalent deep neural network without the biasing applied.

\begin{figure}[h]
  \centering
  \caption{Comparison of expected accuracy against attained accuracy on MNIST-QA with and without Biased Competition. Where Biased Competition is applied, we found that tasks learned barely interfered with one another, whereas when Biased Competition was not applied, there was a strong trend towards interference.}
  \includegraphics[width=0.8\textwidth]{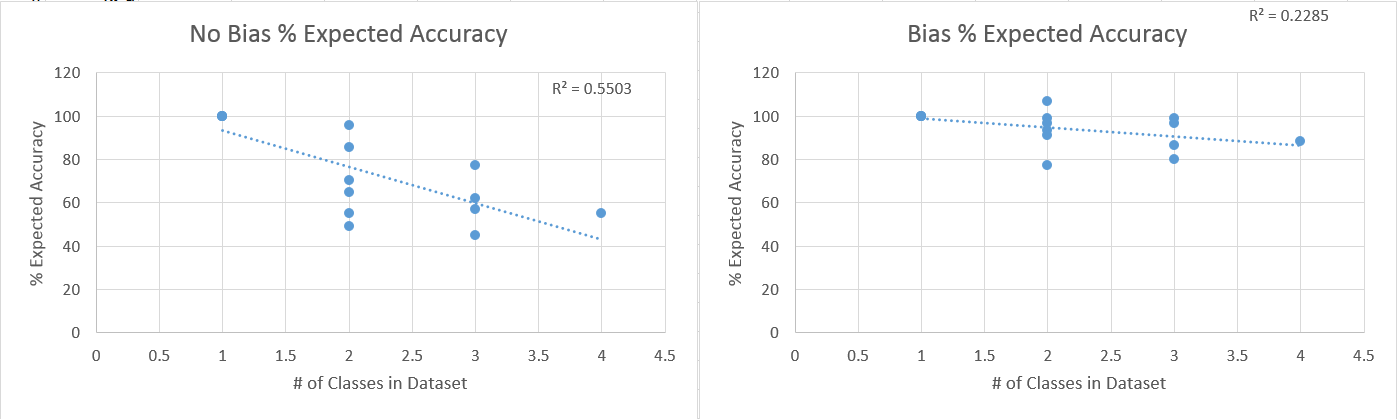}
\end{figure}

We demonstrated that the tasks are separated by the biasing input by performing a cluster analysis on the resultant binary matrices produced. Clusters were found by performing a K-means clustering over the nine tenths of the binary matrices produced by the test set and validated using the remaining tenth. The clusters were well formed and easily separated by task. We verified this by testing the clusters ability to predict question type and found that it was able to do so with 99.6\% accuracy. In addition, we performed a test assigning random labels to ensure the validity of the clusters, which as expected achieved 23.2\% accuracy.  We further calculated the entropy, purity, recall, and f-measure to provide numeric measures of the clusters' cohesion.
\begin{table}[h]
\small
\centering
\caption{Clustering Statistics on MNIST-QA}
\label{clustertable}
\begin{tabular}{|l|l|l|l|l|l|}
\hline
Question Type   & Entropy & Precision & Recall & F-Measure \\ \hline
Count           & 0.0     & 1.0       & 0.9842 & 0.992             \\  \hline
Number          & 0.0     & 1.0       & 0.9842 & 0.992            \\  \hline
Relative Count  & 0.0     & 1.0       & 0.9840 & 0.9919             \\  \hline
Relative Number & 0.081   & 0.984     & 1.0    & 0.992            \\  \hline
\end{tabular}
\end{table}

We can clearly see that the clusters were extremely well formed, indicating a strong separation between tasks. This is further validated by the results of assigning random labels to the binary matrices and reviewing the clustering.

\begin{table}[h]
\small
\centering
\caption{Clustering Statistics on MNIST-QA with random label assignments for each question type}
\label{clusterable2}
\begin{tabular}{|l|l|l|l|l|l|}
\hline
Question Type & Entropy & Precision & Recall & F-Measure  \\ \hline
Count         & 1.374   & 0.216     & 0.0881 & 0.1251     \\ \hline
Number        & 1.385   & 0.228     & 0.0921 & 0.1312     \\  \hline
Relative Count& 1.385   & 0.268     & 0.1029 & 0.1488     \\  \hline
Relative Number&1.383   & 0.267     & 0.1067 & 0.1526     \\  \hline
\end{tabular}
\end{table}
We visualised the binary masks produced by performing t-SNE\citep{maaten2008visualizing}, with colours indicating each question type. Perplexity was set at 50 and the 9000 examples shown were clustered over 5000 iterations. While t-SNE is less than perfect for interpretations of cluster structure, it provides an adequate visualisation which further evidences the separation of question types, and the existence of multiple distinct clusters for each question type hints at the formation of subtasks within those tasks.
\begin{figure}[h]
  \centering
  \caption{t-SNE of binary masks formed into multiple clusters on MNIST-QA}
  \includegraphics[width=0.8\textwidth]{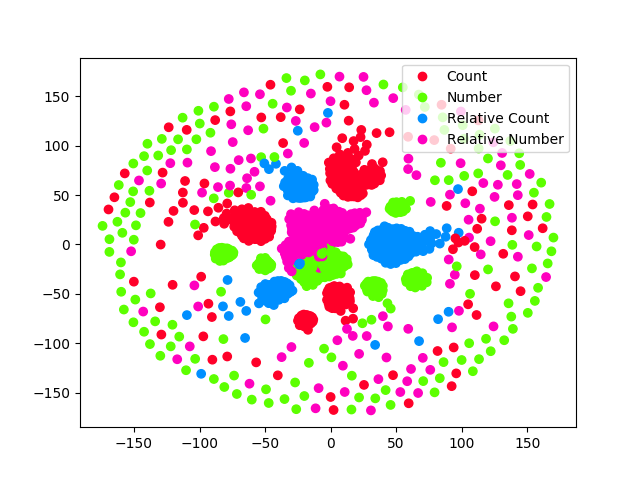}
\end{figure}

We repeated the above experiments for the COCO-QA and DAQUAR datasets and demonstrated that the effect was preserved in the more complex dataset of COCO-QA, though not in the DAQUAR dataset. We discuss possible reasons for this in section 0.5.

\begin{table}[h]
\small
\centering
\caption{Clustering Statistics on DAQUAR}
\label{clusterable3}
\begin{tabular}{|l|l|l|l|l|l|}
\hline
Question Type & Entropy & Precision & Recall & F-Measure  \\ \hline
Colour        & -0.000  & 1.000     & 0.0001 & 0.0003     \\ \hline
Position      & 0.731   & 0.126     & 0.0025 & 0.0050     \\  \hline
Count         & 0.659   & 0.010     & 0.0002 & 0.0004     \\  \hline
Size          & 0.653   & 0.824     & 0.0075 & 0.0148     \\  \hline
\end{tabular}
\end{table}

\begin{table}[h]
\small
\centering
\caption{Clustering Statistics on DAQUAR with random label assignments for each question type}
\label{clusterable4}
\begin{tabular}{|l|l|l|l|l|l|}
\hline
Question Type & Entropy & Precision & Recall & F-Measure  \\ \hline
Colour        & -0.000  & 1.000     & 0.0002 & 0.0004     \\ \hline
Position      & 1.375   & 0.298     & 0.0060 & 0.0117     \\  \hline
Count         & 1.380   & 0.270     & 0.0049 & 0.0096     \\  \hline
Size          & 1.374   & 0.314     & 0.0029 & 0.0057     \\  \hline
\end{tabular}
\end{table}

\begin{figure}[!h]
  \centering
  \caption{t-SNE of binary masks with very weakly formed clusters on DAQUAR}
  \includegraphics[width=0.8\textwidth]{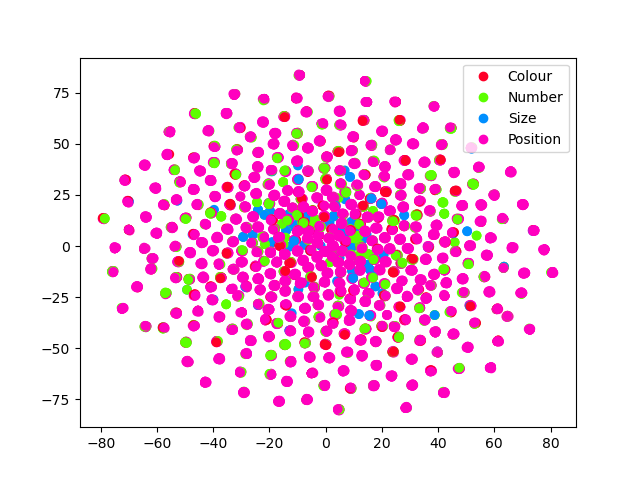}
\end{figure}

\begin{table}[h]
\small
\centering
\caption{Clustering Statistics on COCO-QA}
\label{clusterable5}
\begin{tabular}{|l|l|l|l|l|l|}
\hline
Question Type & Entropy & Precision & Recall & F-Measure  \\ \hline
Colour        & 0.888   & 0.722     & 0.0670 & 0.1226     \\ \hline
Position      & 0.159   & 0.006     & 0.0522 & 0.0110     \\  \hline
Count         & 0.918   & 0.206     & 0.0158 & 0.0294     \\  \hline
Size          & 0.956   & 0.064     & 0.0011 & 0.0021     \\  \hline
\end{tabular}
\end{table}

\begin{table}[h]
\small
\centering
\caption{Clustering Statistics on COCO-QA with random label assignments for each question type}
\label{clusterable6}
\begin{tabular}{|l|l|l|l|l|l|}
\hline
Question Type & Entropy & Precision & Recall & F-Measure  \\ \hline
Colour        & 1.386   & 0.246     & 0.0307 & 0.0546     \\ \hline
Position      & 1.386   & 0.248     & 0.6139 & 0.3535     \\  \hline
Count         & 1.386   & 0.243     & 0.0227 & 0.0417     \\  \hline
Size          & 1.384   & 0.240     & 0.0049 & 0.0097     \\  \hline
\end{tabular}
\end{table}

\begin{figure}[!h]
  \centering
  \caption{t-SNE of binary masks formed into multiple clusters on COCO-QA}
  \includegraphics[width=0.8\textwidth]{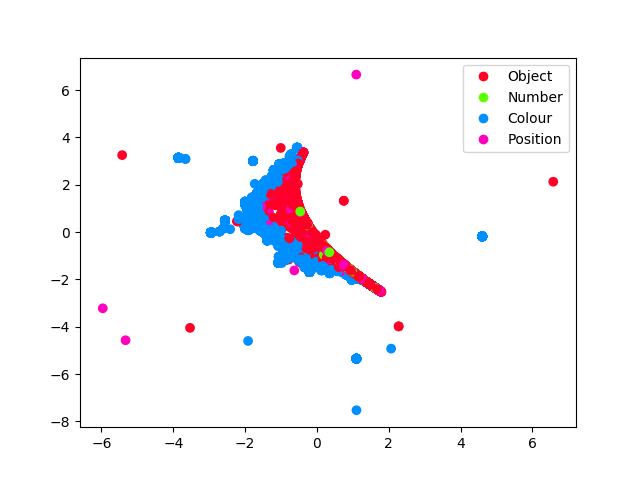}
\end{figure}

\section{Analysis}
We make two claims based on the evidence presented above. Firstly, the Biased Competition model is capable of separating tasks by type, as can clearly be seen by the clustering of the binary matrices produced, which are highly separable and strongly predictive. We can conclude from the existence of these clusters that the network is being separated into distinct subnetworks which solve each task as hypothesised. Secondly, we show that the tasks become independent of one another, with the results presented in \autoref{biastable} showing that the network learned the distinct tasks without negatively impacting the other tasks as indicated by performance matching the expected accuracy.

Although t-SNE clustering is excellent for visualising high dimensional data, the perception of such a visualisation is quite open to interpretation, as slight variations in perplexity can have significant changes in the resulting image with low values allowing the apparent formation of clusters in truly random data. For higher values of perplexity, it is possible that clusters are formed too tightly, eliminating genuine clusters from visibility. For our claims to hold we require the formation of those clusters to be certain, and so we err on the side of caution with a high perplexity value, potentially removing several clusters that may otherwise be visible, if less certain.

The overlap of clusters in the t-SNE visualisation does however imply that the network uses subtasks that are common between tasks, resulting in some amount of overlap between the binary vectors but NOT the clusters in high-dimensional space, therefore the clusters should be separable, but we calculate the overlap by summing the exclusive-or of the binary vectors, and indeed we find this is the case.

For the MNIST-QA dataset, the clustering statistics in \autoref{clustertable} show very strongly separated clusters, with extremely high purity and extremely low entropy. This is not unexpected, as the high-dimensional nature of the data does encourage the formation of extremely pure clusters. However, we are certain that this is not simply an artefact of clustering on high dimensional data, as the results of the clustering with random labelling clearly shows.
With recall also being a strong property of the clustering, we can also confidently state that even complex cases are being accurately categorised, indicating that the Biased Competition method is capable of classifying even the complex questions. However, the network did fail to learn to generate proper binary masks for the DAQUAR dataset, and further analysis of this result indicates that the network is not in fact learning a good representation of the DAQUAR dataset. Instead, the network appears to fall prey to the issue of learning only the most common answers for each question type, achieving an accuracy of 22.2\%. The network does outperform some notable early contributors to the field (7.9\% - Multi-World\citep{malinowski2014nips},21.7\% - Ask Your Neurons\citep{malinowski2015ask}) but fails to achieve state-of-the art, with \citet{yang2016stacked} claiming 29.3\%. This is likely due to the fact that DAQUAR dataset containing relatively few training examples and has a strong presence of a language prior bias, as discussed in \citet{goyal2017making}. Alternatively, as discussed in \citet{ren2015exploring}, it could also be due to the dissimilarity between ImageNet images and those present in the DAQUAR dataset, essentially rendering the visual component meaningless. Fortunately, we find that the larger COCO-QA dataset does not suffer from this issue to the same degree, and a more balanced distribution of answers is generated, achieving an overall accuracy of 54.3\%, which is comparable to the majority of non-attention based Visual Question Answering networks. The clusters formed are not nearly as well structured as those of the simpler MNIST-QA network's, with the k-means algorithm finding 15 clusters which each have a low f-measure score. Though both colour and count question types report a reasonable precision, indicating that they were separated accurately from the other question types, although the corresponding low recall indicates that not all colour and count questions were accurately detected and separated. We attribute this overall lower clustering score to the high error rate of the network itself, as clustering only on correctly answered members of the test set results in much better scores as we show in \autoref{clusterable-acc}. 
\begin{table}[h]
\small
\centering
\caption{Clustering Statistics on COCO-QA with only correctly answered questions included}
\label{clusterable-acc}
\begin{tabular}{|l|l|l|l|l|l|}
\hline
Question Type & Entropy & Precision & Recall & F-Measure  \\ \hline
Colour        & 0.888   & 0.722     & 0.0670 & 0.1226     \\ \hline
Position      & 0.159   & 0.006     & 0.0522 & 0.0110     \\  \hline
Count         & 0.918   & 0.206     & 0.0158 & 0.0294     \\  \hline
Size          & 0.956   & 0.064     & 0.0011 & 0.0021     \\  \hline
\end{tabular}
\end{table}
The collected evidence strongly supports the hypothesis we propose here, both for the formation of distinct subtasks and the significant reduction of inter-task interference.

\section{Conclusion}
The results presented herein lend significant credence to the hypothesis that inter-task catastrophic interference can be strongly suppressed by the action of the Biased Competition model this paper details. The formation of multiple clusters of tasks as demonstrated by the t-SNE visualisation, combined with the strongly separable clusters over the actual task labels found by the k-means clustering show quite strongly that the Biased Competition model acts to separate tasks according to the hypothesised collection of subtasks. This is a significant result as its impact on future work may include applying this technique, or at the very least this understanding of complex tasks, to a variety of neural network architectures.
However, it is important to note that the Biased Competition method has a rather large cost in terms of memory. The binary matrix produced is necessarily of equivalent size to the layer it operates on, thus doubling the memory cost for the affected layers. This is a significant disadvantage, as a major limitation of neural networks is the associated memory requirements and the hardware required to support them. However, this could be improved in future research, which opens up avenues for improving this technique at a later date.
\subsection{Impact}
In a certain subset of problems that can be formatted into a data source and a context vector, this method can significantly augment the ability of networks to learn the component subtasks and prevent interference between distinct tasks in complex datasets. Potential future work may allow this method to be generalised to problem sets without the constraint of a context vector as a separate input. This would allow for this technique to be extended beyond just Visual Question Answering and similar problem sets. Additionally, learning distinct subtasks which are common components of the learned tasks may have applications in the field of transfer learning.
\bibliographystyle{plainnat}
\bibliography{refs}

\end{document}